%% file: main.tex
\documentclass[10pt,twocolumn,letterpaper]{article}
\input{setup/package}
\input{setup/macros}

\input{setup/symbols}

\input{setup/graphicspath}

\newcommand{\mname}{Protein-Mamba}
\usepackage{array}
\usepackage{tabularx}

\usepackage{amsthm}

\newtheorem{problem}{Problem}

\begin{document}

\title{\mname: Biological Mamba Models for Protein Function Prediction}

\author{Bohao Xu$^{1}$, Yingzhou Lu$^{2}$, Yoshitaka Inoue$^3$, Namkyeong Lee$^4$, Tianfan Fu$^5$, Jintai Chen$^{6}$ \\ 
1. Rensselaer Polytechnic Institute, 2. Stanford University, 3. University of Minnesota Twin Cities, \\ 4. KAIST, 5. Rensselaer Polytechnic Institute, 6. University of Illinois Urbana-Champaign }

\maketitle

\begin{abstract} 
Protein function prediction is a pivotal task in drug discovery, significantly impacting the development of effective and safe therapeutics. Traditional machine learning models often struggle with the complexity and variability inherent in predicting protein functions, necessitating more sophisticated approaches. In this work, we introduce Protein-Mamba, a novel two-stage model that leverages both self-supervised learning and fine-tuning to improve protein function prediction. The pre-training stage allows the model to capture general chemical structures and relationships from large, unlabeled datasets, while the fine-tuning stage refines these insights using specific labeled datasets, resulting in superior prediction performance. Our extensive experiments demonstrate that Protein-Mamba achieves competitive performance, compared with a couple of state-of-the-art methods across a range of protein function datasets. This model's ability to effectively utilize both unlabeled and labeled data highlights the potential of self-supervised learning in advancing protein function prediction and offers a promising direction for future research in drug discovery. 
\end{abstract}

\section{Introduction}
Predicting protein function is a fundamental challenge in molecular biology, with profound implications for drug discovery and development. Accurately predicting how proteins function can significantly enhance the efficiency of drug discovery by identifying viable drug candidates earlier in the development process, thus reducing time and costs associated with experimental validation~\cite{wu2022cosbin}. Traditional machine learning models, while useful, often struggle with the complexities and diverse nature of protein data, necessitating the development of more sophisticated approaches.

In recent years, machine learning models have become increasingly important in predicting protein function, offering a way to prioritize potentially desirable proteins without the need for extensive and resource-intensive wet-lab experiments~\cite{huang2020deeppurpose,huang2021therapeutics}. This approach can significantly accelerate the drug discovery process, saving time and resources while improving the chances of identifying viable drug candidates. However, traditional models often struggle with the complexity and variability inherent in protein function prediction, necessitating the development of more sophisticated approaches.

This paper introduces \mname, a two-stage model designed to enhance protein function prediction by leveraging both unlabeled and labeled data through self-supervised learning-based pretraining followed by fine-tuning. By learning from a vast corpus of unlabeled amino acid sequences, during the pretraining stage, \mname~captures underlying chemical structures and relationships, which are then fine-tuned on downstream tasks using labeled datasets. Our results demonstrate that \mname~outperforms several state-of-the-art methods across a range of protein function datasets, highlighting the potential of self-supervised learning in advancing protein function prediction and providing a promising direction for future research in drug discovery.

Our contributions of this paper could be summarized as:
\begin{itemize}[leftmargin=*]
\item We introduce \mname, a two-stage model that combines pre-training and fine-tuning to leverage both unlabeled and labeled data for protein function prediction.
\item \mname~demonstrates competitive performance when compared to a range of state-of-the-art methods across several protein function datasets.
\end{itemize}

\section{Background}

\subsection{Protein}
\label{sec:protein}

Proteins are the building blocks of life. 
Every cell in the human body contains proteins. 
Proteins carry out all the essential functions in the human body. 
Proteins support almost all biological activities. 
For example, stomach enzymes\footnote{Enzymes are proteins that act as biological catalysts by accelerating chemical reactions. } serve to digest food; the movement of myosin on actin is the driving force for muscular contraction; whey proteins strengthen cell anti-oxidation; and antibodies protect us from disease.

\begin{table}[]
\centering
\caption{All the 20 natural amino acids, and their abbreviations and frequencies. 
The frequencies are evaluated on all the processed protein data in the RCSB Protein Data Bank (RCSB PDB) (\url{https://www.rcsb.org/}). 
}
\vspace{1mm}
\resizebox{0.82\columnwidth}{!}{
\begin{tabular}{c|c|c}
\toprule[1pt]
class & abbreviation & percentage (\%)  \\ 
\hline 
Glycine	& Gly / G & 7.6\% \\
Alanine	& Ala / A & 7.7\%   \\
Valine & Val / V & 7.0\%   \\
Leucine & Leu / L & 8.6\%   \\
Isoleucine & Ile / I & 5.5\%  \\
Proline & Pro / P & 4.6\%   \\
Phenylalanine & Phe / F & 3.6\%   \\
Tyrosine & Tyr / Y & 3.1\%  \\
Tryptophan & Trp / W & 1.2\% \\
Serine & Ser / S & 6.7\% \\
Threonine & Thr / T & 5.7\% \\
Cystine & Cys / C & 1.3\% \\
Methionine & Met / M & 2.7\% \\
Asparagine & Asn / N & 4.1\% \\
Glutamine & Gln / Q & 3.9\% \\
Aspartic acid & Asp / D & 5.3\% \\
Glutamic acid & Glu / E & 6.5\% \\
Lysine& Lys / K & 6.3\% \\
Arginine & Arg / R & 5.3\% \\
Histidine & His / H & 3.1\% \\
\bottomrule[1pt]
\end{tabular}
}
\label{table:amino_acid0}
\end{table} 

\begin{figure}
\centering
\includegraphics[width=0.86\columnwidth]{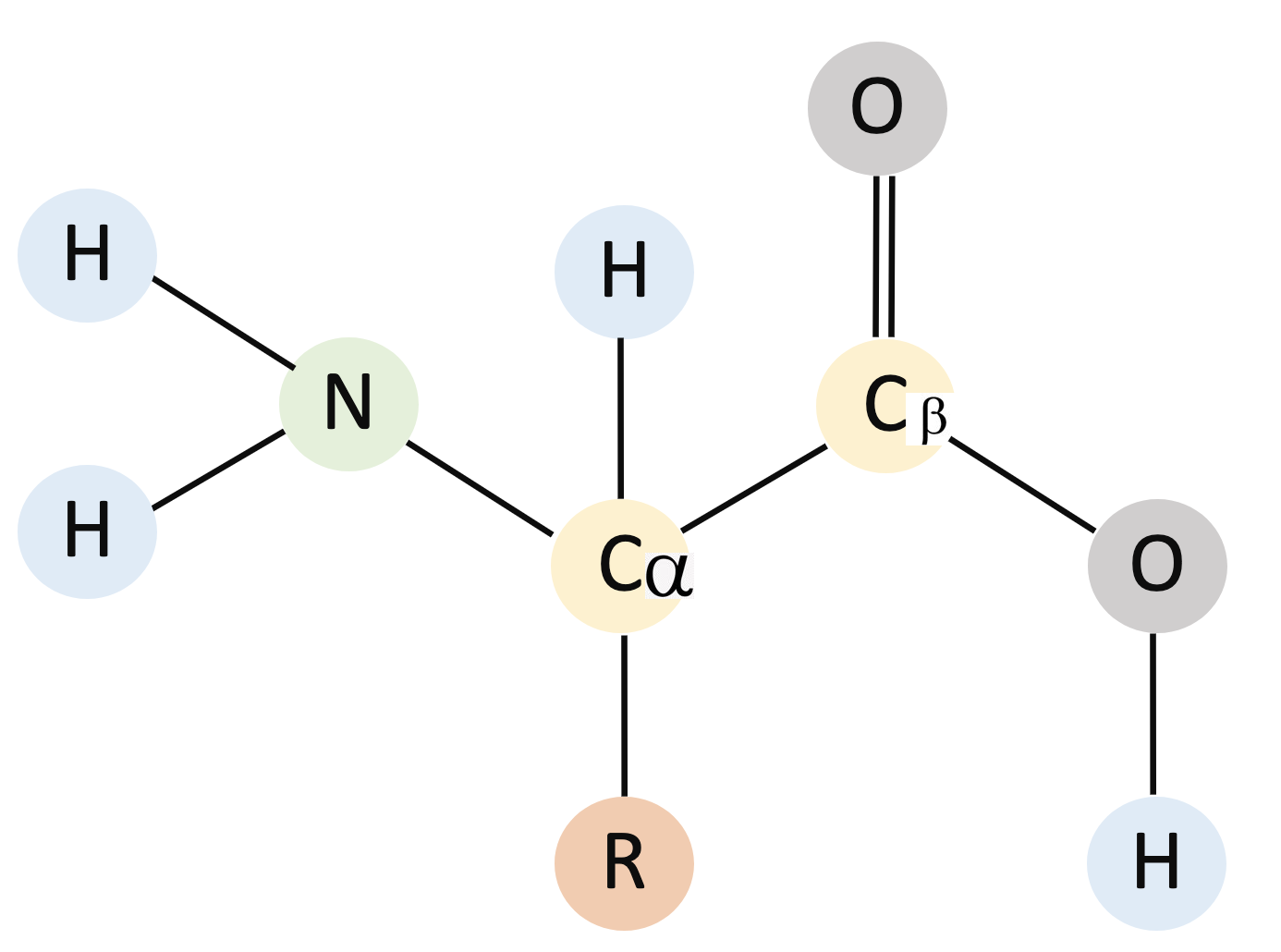}
\caption{Structure of amino acid. 
Amino acids are small organic molecules, where the central carbon atom (denoted $C_{\alpha}$ or alpha-carbon) connects to a carboxyl group (-COOH, C is $C_{\beta}$ or beta-carbon), a hydrogen atom (H), and a variable component called a side chain (denoted residue ``R''). 
The side chain (or amino acid residue) determines the category of amino acids. 
There are 20 kinds of side chains, which means there are 20 kinds of amino acids. 
}
\label{fig:aminoacid}
\end{figure}

\noindent\textbf{Amino acid}. A protein is an amino acid sequence folded into a three-dimensional structure. 
As illustrated in Figure~\ref{fig:aminoacid}, amino acids are small organic molecules that consist of an $\alpha$ (central) carbon atom linked to an amino group (denoted $C_{\alpha}$ in Figure~\ref{fig:aminoacid}), a carboxyl group (-COOH, C is $C_{\beta}$ in Figure~\ref{fig:aminoacid}), a hydrogen atom (H), and a variable component called a side chain (denoted ``R'' in Figure~\ref{fig:aminoacid}). 
The nitrogen (N), alpha carbon ($C_{\alpha}$), and carboxyl carbon ($C_{\beta}$) are in common for all the amino acids and compose the protein backbone. 
The protein backbone holds a protein together and gives it an overall shape. 
The side chain determines the category of amino acids. 
Different amino acids differ in the side chain (R), also referred to as an amino acid residue. 
Multiple amino acids are linked together within a protein by peptide bonds\footnote{Peptide bond refers to the chemical bond connecting two amino acids, which will be described later. }, thereby forming a long chain. 
Amino acids are the basic building blocks of a protein/antibody. 
There are 20 natural amino acids, including Alanine (abbreviated Ala or A), Cysteine (Cys, or C), Glutamine (Gln, or Q), Glycine (Gly, or G), Valine (Val, or V), Leucine (Leu, or L), etc. 
The list of all the 20 natural amino acids, their abbreviations, and frequencies is provided in Table~\ref{table:amino_acid0}. 

\begin{table*}[h]
\caption{Examples of amino acid sequences. A single letter represents one amino acid. The length of an amino acid sequence is typically several hundred, so we only show a part of the entire amino acid sequence. ``1IEP'' is the kinase domain of c-Abl; ``3NY8'' is the human beta2 adrenergic receptor; ``3PBL'' is the human dopamine D3 receptor. }
\vspace{3mm}
\label{table:example_aaseq}
\centering
\begin{tabular}{ll}
\toprule[1pt]
protein ID & amino acid sequence \\ 
\midrule 
1IEP & GAMDPSSPNYDKWEMERTDITMKHKLGGGQY... \\ 
3NY8 & DYKDDDDAMGQPGNGSAFLLAPNRSHAPDHD... \\ 
3PBL & DYKDDDDGAPASLSQLSSHLNYTCGAENSTG... \\ 
% EVQLQQSGAELVKPGASVKLSCTASGFNIKDTYMHWVKQKPEQGLEW... \\ 
% DIVLTQSPAIMSASLGERVTMTCTASSSVSSSNLHWYQQKPGSSPKLWI... \\ 
% EVQLEESGPELVRPGTSVKISCKASGYTFTNYWLGWVKQRPGHGFEWI... \\ 
% QVQLLESGPELKKPGETVKISCKASGYTFTNYGMNWVKQAPGKGLKW... \\ 
% QVQLQQPGAELVKPGASVKLSCKASGYTFTSYWMHWVKQRPGRGLEW...\\ 
\bottomrule[1pt]
\end{tabular}
\end{table*}

A protein is usually represented by four levels of hierarchical structures: 
(1) primary structure (i.e., amino acid sequence), (2) secondary structure, (3) tertiary structure, and (4) quaternary structure. 
They characterize the hierarchical structure of a protein at different levels of complexity, as described below. 
% Suppose we have $N$ amino acids in a protein graph, each amino acid corresponds to a node in the 3D graph structure, a token in the {\it amino acid sequence} and {\it secondary structure sequence}. 
% Next, we describe a few key concepts related to proteins.

\begin{enumerate}
\item \noindent\textbf{Primary structure}. 
The primary structure of a protein is simply the sequence of amino acids in a polypeptide chain. 
% We use $\bfS = (\bfs_1, \cdots, \bfs_{N})$ to denote the amino acid sequence. The length is $N$, $\bfs_i$ represents the $i$-th token (i.e., amino acid) in the sequence. 
We show some examples of amino acid sequences in Table~\ref{table:example_aaseq}. 
\item \noindent\textbf{Secondary structure}. 
The secondary structure of a protein refers to regular, recurring arrangements in the space of adjacent amino acid residues in a polypeptide chain. 
It characterizes the localized conformation of the polypeptide chain. 
% We use $\bfZ = (\bfz_1$, $\cdots$, $\bfz_N)$ to denote the secondary structure sequence. The length is also $N$, $\bfz_i$ represents the kind of secondary structure that $i$-th amino acid belongs to.   
There are a total of 8 categories of secondary structures, e.g., $\alpha$-helix, Turn, Bend, $\pi$-helix, 3-10 helix, Strand, Isolated beta-bridge residue, and None (also called coil, which is not in any above formations)~\cite{narayanan2021machine}. 
Each amino acid belongs to one of these 8 secondary structures. 
\item \noindent\textbf{Tertiary structure}. 
The tertiary structure of a protein refers to the overall three-dimensional arrangement of its polypeptide chain in Euclidean space. 
% We use $\calG = (g_1, \cdots, g_N )$ to denote the 3D graph structure of the protein. There are $N$ nodes in the graph, $g_i$ represents the 3D coordinate of the $i$-th node. 
% $\calG$ is seen as a fully connected 3D graph with different node distances. 
% An amino acid consists of an $\alpha$ (central) carbon atom linked to an amino group, a carboxyl group, a hydrogen atom, and a variable component called a side chain. 
% 3D graph structure is also known as the backbone structure, which belongs to the tertiary structure of a protein. 
\item \noindent\textbf{Quaternary structure}. 
The quaternary structure of a protein describes the relationship of multiple polypeptide chains (or subunits) into a closely packed arrangement. 
Each polypeptide chain has its own primary, secondary, and tertiary structure. 
%The subunits are arranged together by hydrogen bonds and van der Waals forces between nonpolar side chains. It lies in the fourth and highest level of protein structure. 
The proteins with a single polypeptide chain do not have quaternary structures. 
\end{enumerate}

\subsection{Protein Function Prediction}
Protein function prediction is essential in macromolecule drug design because it provides valuable insights into proteins' biological roles and behavior. 
Several key properties of macromolecular drugs impact their qualification as effective pharmaceuticals. 
\begin{itemize}[leftmargin=3mm]
\item Biological activity: Macromolecular drugs must possess the ability to exert the desired effects on target biomolecules or biological processes. They should interact specifically with particular receptors, enzymes, or other biomolecules to modify or regulate their function. The biological activity of a drug directly affects its therapeutic efficacy.
\item Specificity: Macromolecular drugs should exhibit high specificity, meaning they bind selectively to the target molecule or receptor without interacting with non-target molecules. This helps reduce unnecessary side effects and toxicity, enhancing the drug's safety profile.
\item Stability: Macromolecular drugs must exhibit sufficient stability within the body to maintain their structural and functional integrity. They should withstand degradation or inactivation under physiological conditions. Stability is crucial for the drug's longevity and therapeutic effectiveness. 
\item Pharmacokinetic properties: Macromolecular drugs should possess appropriate pharmacokinetic properties in the body, including absorption, distribution, metabolism, and excretion. These properties influence the drug's bioavailability, duration of action, and tissue or organ distribution. 
\item Immunogenicity: Macromolecular drugs should have low immunogenicity, meaning they elicit minimal immune responses within the body. Immune reactions can lead to drug ineffectiveness, allergic reactions, or immune-related adverse events. Thus, minimizing immunogenicity is an important consideration in the design and development of macromolecular drugs. 
\end{itemize}
Understanding a protein's function helps identify potential drug candidates and design drugs that can interact with the protein in a desired manner. For example, the developability of the protein is determined by several factors. Immunogenicity, instability, self-association, high viscosity, poly-specificity, or poor expression can all preclude an antibody from becoming therapeutic. 
If machine learning models could accurately identify these negative characteristics in the early drug discovery process, many high-risk wet lab experiments could be avoided, saving time and money. 

Also, for ease of understanding and quick check, Table~\ref{table:terminology_macromolecule} summarizes most of the important terminologies used in this paper and their descriptions.

% \begin{table}{p{0.27\columnwidth}|p{0.71\columnwidth}}
\begin{table*}
\caption{Glossary of terms used in this paper. }
\label{table:terminology_macromolecule}
\begin{tabular}{l|p{1.7\columnwidth}}
\toprule
{Term} & {Description} \\
\midrule 
Amino acids & Amino acids are building blocks for proteins, which come in 20 different kinds. They can combine to form proteins. \\ 
\hline 
Backbone (protein) & Protein backbone is a continuous chain of a repeated sequence of nitrogen, alpha carbon, and beta carbon, which gives the protein's overall shape.  \\ 
\hline 
Protein function & Protein function refers to the specific role or purpose that a protein serves in a biological system. Proteins have diverse functions, ranging from stability, thermostability, solubility, and biological activity, among many others. The function of a protein is determined by its unique sequence of amino acids, which influences its three-dimensional structure and interactions with other molecules. Understanding protein function is essential for macromolecule drug discovery. 
\\
\hline 
Solubility & Protein solubility is the capacity of a protein to dissolve and remain in solution, playing a crucial role in governing the protein's behavior and functionality across diverse biological and industrial contexts.  \\ 
\hline 
Stability & Protein stability encompasses the capacity of a protein to preserve its three-dimensional structure and functional attributes across diverse environmental conditions, playing a fundamental role in ensuring optimal protein functionality and longevity within biological systems. \\ 
\hline 
Thermostability & Thermostability of a protein refers to its capability to preserve its structural integrity and functional activity even under elevated temperatures. The thermostability of a protein is a critical characteristic that can impact its suitability as a drug candidate. \\ 
\bottomrule
\end{tabular}
\end{table*}

Then, we formulate the protein function prediction problem. 
\begin{problem}[Protein function prediction]
The goal is to predict the protein function $y$ given the protein structure information, typically amino acid sequence. 
Formally, it is defined as 
\begin{equation}
y = f_{\bf{\theta}}(\bf{S}), 
\end{equation}
where $y$ is the label of the problem, which can be either a categorical variable or a continuous value. 
$f_{\bf{\theta}}$ denotes the machine learning model parameterized by $\bf{\theta}$. 
$\bf{S} = (\bf{s}_1, \cdots, \bf{s}_N)$ denotes the amino acid sequence, consisting of $N$ amino acids. 
\end{problem}

\begin{table}[h!]
\centering
\caption{Statistics of protein function prediction datasets. Following the original paper~\cite{xu2022peer}, the training/validation/test splits are carefully designed to make sure all the proteins in the test set are significantly different from the training and validation sets so that we can evaluate the generalization ability of deep learning techniques in real-world settings~\cite{xu2022peer}. ``Seq.~Len.'' represents the average amino acid sequence length and the standard deviation. }
\vspace{1mm}
\label{table:protein_function_dataset}
\resizebox{\columnwidth}{!}{
\begin{tabular}{l|c|c|c|c}
\toprule[1pt]
Function & Dataset & Seq.~Len. & \# train/valid/test & Task \\ 
\hline 
GB1 fitness & FLIP & 378.6(0.9) & 381/43/8,309 & regression \\ 
AAV fitness & FLIP & 1033.0(3.4) & 28,626/3,181/50,776 & regression \\ 
Thermostability & FLIP & 880.6(974.2) & 5,149/643/1,366 & regression \\ 
Fluorescence & Sarkisyan & 343.3(1.3) & 21,446/5,362/27,217 & regression \\ 
Stability & Rocklin & 66.6(5.2) & 53,571/2,512/12,851 & regression \\ 
$\beta$-lactamase activity & Envision & 396.1(0.7) & 4,158/520/520 & regression \\ 
Solubility & DeepSol & 424.1(225.9) & 62,478/6,942/1,999 & classification \\ 
\bottomrule[1pt]
\end{tabular}}
\end{table}

\subsection{Data}
\label{sec:data}
We use machine learning approaches to predict several protein functions, as shown in the following. 
\begin{itemize}[leftmargin=*]
\item \textbf{Solubility}. Protein solubility refers to the ability of a protein to dissolve or remain in a solution. It is an essential property that determines the protein's behavior and functionality in various biological and industrial applications. The solubility of a protein is influenced by several factors, including its amino acid composition, ionic strength, pH, temperature, and the presence of other molecules in the solution. We incorporate the DeepSol dataset for protein solubility prediction. 
\begin{itemize}[leftmargin=*]
\item DeepSol~\cite{khurana2018deepsol}. DeepSol comprises 71,419 proteins with binary labels. The training, validation, and test splits used in our study follow \cite{khurana2018deepsol}, wherein protein sequences exhibiting a sequence identity of 30\% or higher with any sequence in the test set are excluded from the training set. This split assesses the model's capacity to generalize across dissimilar protein sequences. The training, validation, and test set contains 62478, 6942, and 1999 proteins, respectively. 
\end{itemize}
\item \textbf{Thermostability}. The thermostability of a protein refers to its ability to maintain its structural integrity and functional activity at high temperatures. Proteins are composed of intricate three-dimensional structures that are essential for their biological function. However, when exposed to elevated temperatures, proteins can undergo denaturation, which involves the disruption and unfolding of their native structure. This denaturation can lead to the loss of protein function. Thus, protein thermostability is an important characteristic that can influence the potential of a protein to become a drug. We incorporate the FLIP-thermostability dataset for thermostability prediction. 
\begin{itemize}[leftmargin=*] 
\item FLIP-thermostability. The data source is from the FLIP benchmark~\cite{dallago2021flip}. The groundtruth (temperature) is a continuous value. 
The training/validation/test sets consist of 5149, 643, and 1366 data samples, respectively. 
\end{itemize}
\item GB1 fitness. GB1 fitness assesses the fitness values of potential mutants of the GB1 protein~\cite{wu2016adaptation}. This study focuses on the protein G, an immunoglobulin-binding protein, which utilizes the GB1 binding domain to carry out its function. The objective of this investigation is to examine the effects of interactions between mutations and explore ways to enhance the fitness of this crucial functional protein through engineering. 
\begin{itemize}[leftmargin=*]
\item FLIP-GB1. The data source is from the FLIP benchmark~\cite{dallago2021flip}. The groundtruth is the fitness value, a continuous value. The training, validation, and test sets contain 381, 43, and 8309 data points, respectively. 
\end{itemize} 
\item \textbf{AAV fitness}. The manipulation of Adeno-associated virus (AAV) capsid proteins holds significant promise for gene therapy, as it enables the virus to deliver a DNA payload into a specific target cell efficiently. This has garnered considerable interest in the field of gene therapy. AAV fitness evaluates the fitness scores of mutants of VP-1 AAV proteins. 
\begin{itemize}[leftmargin=*]
\item FLIP-AAV. The data source is from the FLIP benchmark~\cite{dallago2021flip}. The groundtruth (fitness score) is a continuous value. The training, validation, and test sets contain 28,626/3,181/50,776 data points, respectively. 
\end{itemize}
\item \textbf{Fluorescence (Flu)}. Protein fluorescence refers to the phenomenon where certain proteins can emit light of a specific wavelength when excited by light of a shorter wavelength. It is a widely used technique to study protein structure, dynamics, interactions, and function. Proteins that exhibit fluorescence typically contain intrinsic fluorophores, such as tryptophan, tyrosine, and phenylalanine residues, which absorb photons and undergo electronic transitions, leading to fluorescent light emission. The emitted fluorescence can be measured and analyzed to provide insights into various aspects of protein behavior, such as folding, conformational changes, binding to other molecules, and enzymatic activity. Protein fluorescence techniques, including fluorescence spectroscopy, fluorescence microscopy, and fluorescence-based assays, offer sensitive and non-invasive tools for studying proteins \textit{in vitro} and \textit{in vivo}. They have wide applications in macromolecule drug discovery. 
\begin{itemize}[leftmargin=*]
\item Sarkisyan~\cite{sarkisyan2016local}. The label is the logarithm of fluorescence intensity. The training and validation sets comprise mutants containing three or fewer mutations, while the test set comprises mutants with four or more mutations. The training, validation, and test sets contain 21446, 5362, and 27217 proteins, respectively. 
\end{itemize}
\item \textbf{$\beta$-lactamase activity}. This task focuses on investigating the enhancement of activity for $\beta$-lactamase, the most prevalent enzyme responsible for conferring beta-lactam antibiotic resistance to gram-negative bacteria through single mutations.
\begin{itemize}[leftmargin=*]
\item Envision~\cite{gray2018quantitative}. The groundtruth represents the experimentally tested fitness score, which quantifies the scaled mutation effect for each mutant. The training, validation, and test sets contain 4158, 520, and 520 proteins, respectively. 
\end{itemize}
\item \textbf{Stability (Sta)}. Protein stability refers to the ability of a protein to maintain its three-dimensional structure and functional properties under various environmental conditions. It is crucial for the proper functioning and longevity of proteins in biological systems. The stability of a protein is determined by its ability to resist denaturation, aggregation, and degradation.
\begin{itemize}[leftmargin=*]
\item Rocklin dataset was collected by~\cite{rocklin2017global} to measure protein stability experimentally. The proteins obtained from four rounds of experimental design are utilized for training and validation purposes. Then, the top candidates containing single mutations are employed for testing. This evaluation assesses the model's generalization ability, as it is trained on data with multiple mutations and then applied to identify the top candidates with single mutations. The training, validation, and test sets contain 53,571, 2,512, and 12,851 proteins. 
\end{itemize}
% \item \textbf{Developability}. Protein developability refers to the set of properties and characteristics that make a protein molecule suitable for development as a therapeutic agent. It encompasses various factors that determine the feasibility, efficacy, safety, and manufacturability of a protein for practical applications. 
\end{itemize}
The statistics of all the datasets are described in Table~\ref{table:protein_function_dataset}.

\section{Method: \mname}

\subsection{Overview} 
The \mname~model utilizes a two-stage approach involving pre-training and fine-tuning to improve protein function prediction by efficiently leveraging both unlabeled and labeled data. 
We first describe the basic Mamba model in Section~\ref{sec:mamba}. The pretraining and finetuning steps are described in Section~\ref{sec:pretrain} and Section~\ref{sec:finetune}, respectively.

\subsection{Model Backbone: Mamba}
\label{sec:mamba} 

Mamba~\cite{gu2023mamba} is a version of the Structured State Space Sequence (S4) model that is specifically designed to handle long-term patterns in sequential data. Unlike traditional models, Mamba is particularly good at tasks like time-series analysis and natural language processing because it can capture both short-term and long-term patterns within sequences. It uses state space models to keep track of and update hidden states over long sequences, which helps in accurately modeling complex time-based data. Mamba’s design also allows for efficient parallel processing, making it scalable for large datasets. This makes it especially useful in areas where understanding long-term dependencies is important. Mamba has been successfully applied in various fields, including natural language processing~\cite{waleffe2024empirical,yue2024biomamba}, computer vision~\cite{yu2024mambaout,wang2024large}, EEG analysis~\cite{xu2024mambacapsule}, and drug discovery~\cite{xu2024smilesmamba}.

Transformers~\cite{vaswani2017attention} and Mamba are both powerful models for handling sequential data, but they employ different approaches and dominate in distinct areas. Transformers utilize self-attention mechanisms to capture dependencies within sequences, making them particularly effective for tasks like natural language processing and machine translation, where they can model relationships between all sequence elements simultaneously. However, they can face challenges with very long sequences due to their high computational complexity. On the other hand, Mamba, which is built on the Structured State Space Sequence (S4) model, is optimized for efficiently managing long-range dependencies by using state space models that maintain and update hidden states over extended sequences. This design makes Mamba especially suitable for tasks like time-series analysis, where capturing long-term temporal patterns is essential. While Transformers offer versatility and robust performance across a wide range of tasks, Mamba is experienced at handling long sequences and long-range dependencies.

\subsection{Pretraining on Large Unlabelled Data}
\label{sec:pretrain}
Pretraining is crucial as it enables a model to learn general features and patterns from large datasets, which can later be fine-tuned for specific downstream tasks using smaller labeled datasets. This approach greatly enhances the model's performance, reduces the need for extensive labeled data, and speeds up training for downstream tasks by beginning with a well-initialized model rather than starting from scratch. 

During the pre-training stage, the model is trained on a large corpus of unlabeled protein data, such as amino acid sequences, to learn the underlying biological structures and relationships. This stage enables the model to develop a comprehensive representation of protein features without relying on explicit labels, capturing crucial patterns and dependencies within the data. The Mamba model, which is autoregressive, uses next-step prediction as its pretraining objective. Since the dataset lacks any labels related to protein function, the pretrained Mamba model remains function-agnostic. 

\noindent\textbf{RCSB}. We use protein amino acid sequence for pretraining. Specifically, we download all the protein data in pdb format from \url{https://www.rcsb.org/}. 
The Protein Data Bank (PDB) file format is a textual format used to describe the three-dimensional structures of molecules stored in the Protein Data Bank. This format includes detailed descriptions and annotations of protein and nucleic acid structures, such as atomic coordinates, secondary structure assignments, and atomic connectivity. A comprehensive list of all amino acids, their secondary structures, and their frequencies is available in the supplementary materials. Our dataset includes 27,043 proteins, each with a single chain. We use it to pretrain a function-agnostic Mamba model.

\subsection{Fine-tuning on Downstream Tasks}
\label{sec:finetune}

After pre-training, the model is fine-tuned using a smaller, labeled dataset tailored to the target task, such as predicting protein functions like solubility, stability, and thermostability. Fine-tuning involves adjusting the pre-trained model's parameters to optimize its performance on the specific task, leveraging the labeled data to refine and improve its predictions. This two-stage process greatly enhances the model's ability to predict protein functions by combining the broad generalization capabilities acquired during pre-training with the task-specific insights obtained through fine-tuning. 

By leveraging both unlabeled and labeled data, the \mname~model delivers superior prediction performance, establishing itself as a powerful tool in drug discovery and other applications that demand precise protein function prediction. This approach reduces the dependence on large quantities of labeled data, which are often scarce and expensive to acquire.

\section{Experiment}
\label{sec:experiment}
In this section, we provide a detailed overview of the empirical studies, covering baseline methods, evaluation metrics, implementation details, experimental results, and their subsequent analysis.

\subsection{Experimental Setup}
\paragraph{Evaluation Metrics.}
Protein function prediction can be categorized into two machine learning tasks (classification and regression) based on the groundtruth. 
For classification tasks (mostly binary classification), we use accuracy as metric. 
On the other hand, for regression tasks, we select Spearman as the evaluation metric: 
{Spearman's rank correlation coefficient (Spearman)} is the Pearson correlation coefficient between the rank variables. Higher values indicate better performance. It is used when a trend (ranking) is more important than the absolute error. 

\paragraph{Baseline Methods.}
The key challenge of protein function prediction is to obtain a powerful representation of the protein amino acid sequence. 
We cover various neural network architectures, as shown in the following. 
\begin{enumerate}
% \item \textbf{Multiple layer perceptron (MLP)}. It uses xxx as the input feature. The fingerprint vectors of the reactants and product are concatenated and fed into a multiple-layer perceptron (MLP). The MLP has five layers. The dimensions of the hidden state are 1024, 1024, 512, 128, and 32, respectively. Details of MLP can be referred to Section~\ref{sec:mlp}. 
\item \textbf{One-dimensional Convolutional neural network (1D-CNN)}. 1D-CNN takes amino acid sequences as the input. 1D-CNN has four layers; the number of filters for the four layers is 32, 64, 96, and 128, respectively. The kernel sizes are 6, 8, 10, and 12, respectively. The convolutional layer is followed by a one-layer MLP to produce the prediction, which is a scalar. 
\item \textbf{Long short-term memory (LSTM)}. We use the bidirectional three-layer LSTM to represent the amino acid sequence. Bidirectional LSTM scans the amino acid sequences based on forward and reverse order. 
The hidden variables of the last hidden state from both forward and reverse order are concatenated as the output of bidirectional LSTM, which is followed by a two-layer MLP and makes the prediction. In LSTM, the dimension of the hidden state is set to 100. 
\item \textbf{Transformer}. We use the transformer encoder~\cite{vaswani2017attention} to represent the amino acid sequence. Three layers of transformer architectures are stacked. The dimension of embedding in the transformer is set to 100. The number of attention heads is set to 8. 
\item \textbf{ResNet+1D-CNN (ResNet)}. ResNet+1D-CNN utilizes the residual network~\cite{he2016deep} to enhance 1D-CNN to alleviate the gradient vanishment issue. Specifically, the residual network skips some hidden-layer connections and jumps over some hidden layers in CNN. The other setup is the same as the 1D-CNN mentioned above. 
\item \textbf{Protein BERT (ProtBERT)~\cite{brandes2022proteinbert}}. Bidirectional Encoder Representations from Transformers (BERT) is a transformer-based neural network architecture pretrained on unlabeled sequence data. 
Here, BERT~\cite{devlin2018bert} is pretrained on a large number of unlabeled amino acid sequences and then finetunes on labeled training datasets. The sizes of the token and positional embeddings are both 1024. It stacks 30 transformer blocks, and the number of attention heads is 16. 
% \item \textbf{ESM}. Evolutionary Scale Modeling (ESM) is a pretrained language model for proteins (pretrained on large-scale amino acid sequence databases). It utilizes 33 Transformer blocks with 20 attention heads for each block. The hidden dimension is set to 1280. 
\end{enumerate}

\begin{table*}[h!]
\centering
\caption{
Experimental results of protein function prediction on various datasets. On each task, we highlight the best method. For all the tasks, higher values indicate better performance (denoted $\uparrow$). }
\vspace{1mm}
\resizebox{2\columnwidth}{!}{
\begin{tabular}{lccccccccc}
\toprule[1pt]
Method & GB1 & AAV & Thermo & Flu & Stab & $\beta$-lac & Solubi \\
\hline
% DDE & 0.445(0.023) & 0.649(0.012) & 0.349(0.007) & 0.638(0.003) & 0.652(0.033) & 0.623(0.019) & 59.77(1.21) \\
% Moran & 0.069(0.003) & 0.437(0.008) & 0.331(0.003) & 0.400(0.001) & 0.322(0.011) & 0.375(0.008) & 57.73(1.33) \\
Metric & Spearman ($\uparrow$) & Spearman ($\uparrow$) & Spearman ($\uparrow$) & Spearman ($\uparrow$) & Spearman ($\uparrow$) & Spearman ($\uparrow$) & accuracy (\%) ($\uparrow$) \\ 
\hline 
LSTM & -0.002(0.003) & 0.125(0.025) & 0.564(0.007) & 0.494(0.071) & 0.533(0.101) & 0.139(0.051) & \bf  70.18(0.63) \\
Transformer & 0.271(0.020) & 0.681(0.013) & 0.545(0.031) & 0.643(0.005) & 0.649(0.056) & 0.261(0.015) & 70.12(0.31) \\
CNN & 0.502(0.007) & 0.746(0.003) & 0.494(0.021) & {0.682(0.002)} & 0.637(0.010) & 0.781(0.011) & 64.43(0.25) \\
ResNet & 0.133(0.095) & 0.739(0.013) & 0.528(0.009) & 0.636(0.021) & 0.126(0.094) & 0.152(0.029) & 67.33(1.46) \\
ProtBERT & 0.634(0.047) & \bf 0.794(0.014) & 0.660(0.009) & 0.679(0.001) & \textbf{0.771(0.020)} & 0.731(0.226) & 68.15(0.92) \\
% ProtBERT* & 0.123(0.012) & 0.209(0.001) & 0.562(0.001) & 0.339(0.003) & 0.697(0.013) & 0.616(0.002) & 59.17(0.21) \\
% ESM & {0.704(0.018)} & \textbf{0.821(0.010)} & {0.669(0.028)} & 0.679(0.002) & 0.694(0.073) & \textbf{0.839(0.053)} & \textbf{70.23(0.75)} \\
\mname & \bf 0.706(0.013) & 0.754(0.008) & \bf 0.674(0.002) & \bf 0.683(0.002) & 0.753(0.010) & \bf 0.788(0.009) & 68.25(0.40) \\ 
\bottomrule[1pt]
\end{tabular}}
\label{table:biologic_property_prediction}
\end{table*}

\paragraph{Implementation Details. }
All the experiments are conducted on an NVIDIA GeForce RTX 3090 GPU. \mname is implemented in Python 3.8 and PyTorch 1.9.0. The Adam~\cite{kingma2014adam} is used as numerical optimizer with initial learning rate 1e-3. In pretraining phase, the maximal epoch number is set to 100. In finetuning process, the maximal epoch number is set to 50. We use early stop strategy to save the training time and computational resources and avoid overfitting. The \mname~architecture consists of 8 layers, the hidden state dimension is set to 300, while the number of attention heads is set to 8.

\subsection{Results \& analysis.}

The results for absorption, distribution, metabolism, excretion, and toxicity property prediction are reported in Table~\ref{table:biologic_property_prediction}. 
By carefully comparing all the results, we draw a couple of conclusions as follows: 
\begin{itemize}
\item First, the proposed \mname~model exhibits competitive performance in all the protein function tasks. Concretely, compared with five cutting-edge machine learning models, it achieves the highest score in four tasks and top-2 performance in six tasks among all seven various protein function prediction tasks. 
\item Secondly, self-supervised learning-based pretraining strategies have shown to be highly effective. Models such as the proposed \mname~and ProtBERT~\cite{brandes2022proteinbert} stand out by utilizing self-supervised learning to derive valuable insights from unlabeled data. These approaches underscore the potential of self-supervised learning as a promising avenue for future research, highlighting its substantial impact on improving model performance in protein function prediction. 
\item Third, no single method performs best across all tasks, as results can differ based on the type of features and the specific task. This difference comes from the unique information captured by various protein representations and machine learning models. Therefore, combining these different feature types could lead to even better model performance. 
\end{itemize}

\section{Conclusion}
In this paper, we presented \mname, a novel two-stage model for protein function prediction that leverages both unlabeled and labeled data. By integrating self-supervised pretraining with fine-tuning, \mname~effectively captures the intricate relationships within protein data. Our extensive experiments demonstrated that \mname~outperforms several state-of-the-art models across a variety of protein function prediction tasks, underscoring the effectiveness of self-supervised learning in this domain. By reducing the dependency on large labeled datasets, \mname~not only improves prediction accuracy but also opens new avenues for research in drug discovery, potentially accelerating the identification and development of safe and effective drug candidates. The success of \mname~highlights the critical role of advanced machine learning techniques in tackling the complex challenges inherent in drug discovery and development.

Future research can be pursued in the following three areas: (1) In early-stage clinical trials, precise protein function profiling is crucial as it helps researchers understand how a drug is absorbed, distributed, metabolized by enzymes, excreted, and whether it poses any toxic risks. This detailed information enables the identification of potential safety concerns before large-scale trials, thereby preventing costly failures at later stages~\cite{yue2024ct,yue2024trialenroll,lu2024uncertainty,chen2024uncertainty}; (2) Integration of protein function data with multi-omics: Combining protein function data with genetic, transcriptomic, and metabolic information allows researchers to gain deeper insights into how these variations influence drug behavior and response across different individuals or populations~\cite{lu2019integrated,lu2021cot,lu2022cot,chen2021data}. This integration facilitates the identification of biomarkers for predicting drug efficacy and toxicity, supports the development of more effective and personalized therapies, and helps minimize adverse drug reactions~\cite{fu2024ddn3,zhang2021ddn2}. 

{\small
\bibliographystyle{plain}
\bibliography{main.bib}
}

\end{document}

%% file: setup/package.tex
\usepackage{graphicx}
\usepackage{subcaption}
\usepackage{float}
\usepackage[justification=raggedright]{caption}	% makes captions ragged right - thanks to Bryce Lobdell
\usepackage{lscape}                                         % Useful for wide tables or figures.

\usepackage[lined,ruled,linesnumbered]{algorithm2e}

\usepackage{booktabs}                   % Publication quality tables
\usepackage{multirow}

\usepackage{paralist}
\usepackage{enumitem}

\usepackage{bm}                          % Make bold, italic math symbols
\usepackage{epsfig}                      % for figures
\usepackage{graphicx}                  % another package that works for figures
\usepackage{times}
\usepackage{mathptmx}
\usepackage{mathtools}
\usepackage{amssymb,amsmath}   % Short math guide for LaTeX ftp://ftp.ams.org/pub/tex/doc/amsmath/short-math-guide.pdf

\usepackage{units}
\usepackage{color}

\usepackage{comment}

\usepackage{url}  % Hyphenation of URLs.
\usepackage[pagebackref=true,breaklinks=true,letterpaper=true,colorlinks,bookmarks=false]{hyperref}

\usepackage{xspace}
\usepackage[table]{xcolor}
\usepackage{setspace}

%% file: setup/macros.tex
\newlength\paramargin
\newlength\figmargin
\newlength\secmargin

\long\def\ignorethis#1{}

%% file: setup/symbols.tex
\def\xi{\mathbf{x}_i}

%% file: setup/graphicspath.tex
\graphicspath{{figure}, {example}}

%% file: main.bbl
\begin{thebibliography}{10}

\bibitem{brandes2022proteinbert}
Nadav Brandes, Dan Ofer, Yam Peleg, Nadav Rappoport, and Michal Linial.
\newblock {ProteinBERT}: A universal deep-learning model of protein sequence and function.
\newblock {\em Bioinformatics}, 38(8):2102--2110, 2022.

\bibitem{lu2019integrated}
Yi-Tan Chang, Eric~P Hoffman, Guoqiang Yu, David~M Herrington, Robert Clarke, Chiung-Ting Wu, Lulu Chen, and Yue Wang.
\newblock Integrated identification of disease specific pathways using multi-omics data.
\newblock {\em bioRxiv}, page 666065, 2019.

\bibitem{chen2021data}
Lulu Chen, Yingzhou Lu, Chiung-Ting Wu, Robert Clarke, Guoqiang Yu, Jennifer~E Van~Eyk, David~M Herrington, and Yue Wang.
\newblock Data-driven detection of subtype-specific differentially expressed genes.
\newblock {\em Scientific reports}, 11(1):332, 2021.

\bibitem{chen2024uncertainty}
Tianyi Chen, Nan Hao, Yingzhou Lu, and Capucine Van~Rechem.
\newblock Uncertainty quantification on clinical trial outcome prediction.
\newblock {\em arXiv preprint arXiv:2401.03482}, 2024.

\bibitem{dallago2021flip}
Christian Dallago, Jody Mou, Kadina~E Johnston, Bruce~J Wittmann, Nicholas Bhattacharya, Samuel Goldman, Ali Madani, and Kevin~K Yang.
\newblock Flip: Benchmark tasks in fitness landscape inference for proteins.
\newblock {\em bioRxiv}, pages 2021--11, 2021.

\bibitem{devlin2018bert}
Jacob Devlin, Ming{-}Wei Chang, Kenton Lee, and Kristina Toutanova.
\newblock {BERT:} pre-training of deep bidirectional transformers for language understanding.
\newblock In {\em Proceedings of the 2019 Conference of the North American Chapter of the Association for Computational Linguistics: Human Language Technologies, {NAACL-HLT} 2019.}, pages 4171--4186. Association for Computational Linguistics, 2019.

\bibitem{fu2024ddn3}
Yi~Fu, Yingzhou Lu, Yizhi Wang, Bai Zhang, Zhen Zhang, Guoqiang Yu, Chunyu Liu, Robert Clarke, David~M Herrington, and Yue Wang.
\newblock Ddn3. 0: Determining significant rewiring of biological network structure with differential dependency networks.
\newblock {\em Bioinformatics}, page btae376, 2024.

\bibitem{gray2018quantitative}
Vanessa~E Gray, Ronald~J Hause, Jens Luebeck, Jay Shendure, and Douglas~M Fowler.
\newblock Quantitative missense variant effect prediction using large-scale mutagenesis data.
\newblock {\em Cell systems}, 6(1):116--124, 2018.

\bibitem{gu2023mamba}
Albert Gu and Tri Dao.
\newblock Mamba: Linear-time sequence modeling with selective state spaces.
\newblock {\em arXiv preprint arXiv:2312.00752}, 2023.

\bibitem{he2016deep}
Kaiming He, Xiangyu Zhang, Shaoqing Ren, and Jian Sun.
\newblock Deep residual learning for image recognition.
\newblock In {\em CVPR}, 2016.

\bibitem{huang2021therapeutics}
Kexin Huang, Tianfan Fu, Wenhao Gao, Yue Zhao, Yusuf Roohani, Jure Leskovec, Connor~W Coley, Cao Xiao, Jimeng Sun, and Marinka Zitnik.
\newblock Therapeutics data commons: machine learning datasets and tasks for therapeutics.
\newblock {\em NeurIPS Track Datasets and Benchmarks}, 2021.

\bibitem{huang2020deeppurpose}
Kexin Huang, Tianfan Fu, Lucas~M Glass, Marinka Zitnik, Cao Xiao, and Jimeng Sun.
\newblock {DeepPurpose}: a deep learning library for drug--target interaction prediction.
\newblock {\em Bioinformatics}, 36(22-23):5545--5547, 2020.

\bibitem{khurana2018deepsol}
Sameer Khurana, Reda Rawi, Khalid Kunji, Gwo-Yu Chuang, Halima Bensmail, and Raghvendra Mall.
\newblock {DeepSol}: a deep learning framework for sequence-based protein solubility prediction.
\newblock {\em Bioinformatics}, 34(15):2605--2613, 2018.

\bibitem{kingma2014adam}
Diederik~P Kingma and Jimmy Ba.
\newblock Adam: A method for stochastic optimization.
\newblock {\em International Conference on Learning Representations}, 2014.

\bibitem{lu2024uncertainty}
Yingzhou Lu, Tianyi Chen, Nan Hao, Capucine Van~Rechem, Jintai Chen, and Tianfan Fu.
\newblock Uncertainty quantification and interpretability for clinical trial approval prediction.
\newblock {\em Health Data Science}, 4:0126, 2024.

\bibitem{lu2021cot}
Yingzhou Lu, Chiung-Ting Wu, Sarah~J Parker, Lulu Chen, Georgia Saylor, Jennifer~E Van~Eyk, David~M Herrington, and Yue Wang.
\newblock {COT}: an efficient python tool for detecting marker genes among many subtypes.
\newblock {\em bioRxiv}, pages 2021--01, 2021.

\bibitem{lu2022cot}
Yingzhou Lu, Chiung-Ting Wu, Sarah~J Parker, Zuolin Cheng, Georgia Saylor, Jennifer~E Van~Eyk, Guoqiang Yu, Robert Clarke, David~M Herrington, and Yue Wang.
\newblock {COT}: an efficient and accurate method for detecting marker genes among many subtypes.
\newblock {\em Bioinformatics Advances}, 2(1):vbac037, 2022.

\bibitem{narayanan2021machine}
Harini Narayanan, Fabian Dingfelder, Alessandro Butt{\'e}, Nikolai Lorenzen, Michael Sokolov, and Paolo Arosio.
\newblock Machine learning for biologics: opportunities for protein engineering, developability, and formulation.
\newblock {\em Trends in pharmacological sciences}, 2021.

\bibitem{rocklin2017global}
Gabriel~J Rocklin, Tamuka~M Chidyausiku, Inna Goreshnik, Alex Ford, Scott Houliston, Alexander Lemak, Lauren Carter, Rashmi Ravichandran, Vikram~K Mulligan, Aaron Chevalier, et~al.
\newblock Global analysis of protein folding using massively parallel design, synthesis, and testing.
\newblock {\em Science}, 357(6347):168--175, 2017.

\bibitem{sarkisyan2016local}
Karen~S Sarkisyan, Dmitry~A Bolotin, Margarita~V Meer, Dinara~R Usmanova, Alexander~S Mishin, George~V Sharonov, Dmitry~N Ivankov, Nina~G Bozhanova, Mikhail~S Baranov, Onuralp Soylemez, et~al.
\newblock Local fitness landscape of the green fluorescent protein.
\newblock {\em Nature}, 533(7603):397--401, 2016.

\bibitem{vaswani2017attention}
Ashish Vaswani, Noam Shazeer, Niki Parmar, Jakob Uszkoreit, Llion Jones, Aidan~N Gomez, {\L}ukasz Kaiser, and Illia Polosukhin.
\newblock Attention is all you need.
\newblock In {\em Advances in neural information processing systems}, pages 5998--6008, 2017.

\bibitem{waleffe2024empirical}
Roger Waleffe, Wonmin Byeon, Duncan Riach, Brandon Norick, Vijay Korthikanti, Tri Dao, Albert Gu, Ali Hatamizadeh, Sudhakar Singh, Deepak Narayanan, et~al.
\newblock An empirical study of mamba-based language models.
\newblock {\em arXiv preprint arXiv:2406.07887}, 2024.

\bibitem{wang2024large}
Jinhong Wang, Jintai Chen, Danny Chen, and Jian Wu.
\newblock Large window-based mamba unet for medical image segmentation: Beyond convolution and self-attention.
\newblock {\em arXiv preprint arXiv:2403.07332}, 2024.

\bibitem{wu2022cosbin}
Chiung-Ting Wu, Minjie Shen, Dongping Du, Zuolin Cheng, Sarah~J Parker, Jennifer~E Van~Eyk, Guoqiang Yu, Robert Clarke, David~M Herrington, et~al.
\newblock Cosbin: cosine score-based iterative normalization of biologically diverse samples.
\newblock {\em Bioinformatics Advances}, 2(1):vbac076, 2022.

\bibitem{wu2016adaptation}
Nicholas~C Wu, Lei Dai, C~Anders Olson, James~O Lloyd-Smith, and Ren Sun.
\newblock Adaptation in protein fitness landscapes is facilitated by indirect paths.
\newblock {\em Elife}, 5:e16965, 2016.

\bibitem{xu2024smilesmamba}
Bohao Xu, Yingzhou Lu, Chenhao Li, Ling Yue, Xiao Wang, Nan Hao, Tianfan Fu, and Jintai Chen.
\newblock Smiles-mamba: Chemical mamba foundation models for drug admet prediction.
\newblock {\em arXiv preprint arXiv:2408.02600}, 2024.

\bibitem{xu2022peer}
Minghao Xu, Zuobai Zhang, Jiarui Lu, Zhaocheng Zhu, Yangtian Zhang, Ma~Chang, Runcheng Liu, and Jian Tang.
\newblock Peer: a comprehensive and multi-task benchmark for protein sequence understanding.
\newblock {\em Advances in Neural Information Processing Systems}, 35:35156--35173, 2022.

\bibitem{xu2024mambacapsule}
Yinlong Xu, Xiaoqiang Liu, Zitai Kong, Yixuan Wu, Yue Wang, Yingzhou Lu, Honghao Gao, Jian Wu, and Hongxia Xu.
\newblock Mambacapsule: Towards transparent cardiac disease diagnosis with electrocardiography using mamba capsule network.
\newblock {\em arXiv preprint arXiv:2407.20893}, 2024.

\bibitem{yu2024mambaout}
Weihao Yu and Xinchao Wang.
\newblock Mambaout: Do we really need mamba for vision?
\newblock {\em arXiv preprint arXiv:2405.07992}, 2024.

\bibitem{yue2024ct}
Ling Yue and Tianfan Fu.
\newblock Ct-agent: Clinical trial multi-agent with large language model-based reasoning.
\newblock {\em arXiv preprint arXiv:2404.14777}, 2024.

\bibitem{yue2024trialenroll}
Ling Yue, Sixue Xing, Jintai Chen, and Tianfan Fu.
\newblock Trialenroll: Predicting clinical trial enrollment success with deep \& cross network and large language models.
\newblock {\em arXiv preprint arXiv:2407.13115}, 2024.

\bibitem{yue2024biomamba}
Ling Yue, Sixue Xing, Yingzhou Lu, and Tianfan Fu.
\newblock Biomamba: A pre-trained biomedical language representation model leveraging mamba.
\newblock {\em arXiv preprint arXiv:2408.02600}, 2024.

\bibitem{zhang2021ddn2}
Bai Zhang, Yi~Fu, Yingzhou Lu, Zhen Zhang, Robert Clarke, Jennifer~E Van~Eyk, David~M Herrington, and Yue Wang.
\newblock {DDN}2.0: R and python packages for differential dependency network analysis of biological systems.
\newblock {\em bioRxiv}, pages 2021--04, 2021.

\end{thebibliography}
